\documentclass{article}
\usepackage{preprint,times}

\usepackage{amsmath,amsfonts,bm}

\usepackage{booktabs}
\usepackage{graphicx}
\usepackage{algorithm}
\usepackage{algorithmic}

\usepackage{multirow}
\usepackage{wrapfig}
\usepackage{capt-of}
\usepackage{float}
\usepackage{amssymb}
\usepackage[mode=text,detect-weight=true]{siunitx}

\usepackage{hyperref}
\usepackage{url}

\title{WAM-Cache: Staleness-Bounded KV Reuse for Efficient World Action Models}

\author{
  Kai Ding$^{1}$ \quad Yang He$^{2}$ \quad Ruijie Quan$^{1}$ \quad Yi Yang$^{1}$ \\
  $^{1}$Zhejiang University \quad $^{2}$IAIC, Agency for Science, Technology and Research, Singapore \\
  \texttt{dingkai@zju.edu.cn, he\_yang@a-star.edu.sg, quanruijie@zju.edu.cn} \\
  \texttt{yangyics@zju.edu.cn} \\
  Project page: \url{https://dingkai0302.github.io/wam-cache/}
}

\begin{document}

\maketitle

\begin{abstract}
World Action Models (WAMs) enable generalist robot manipulation by conditioning an action expert on representations from a pretrained video Diffusion Transformer (DiT).
In closed-loop control, the video DiT runs at every chunk to encode the current observation into layerwise key--value (KV) pairs that the action expert queries.
This prefill dominates the per-chunk computational cost, yet existing training-free accelerations leave it fully dense.
We present \textbf{WAM-Cache}, a training-free framework that retains layerwise key--value representations across chunks and recomputes only a sparse \emph{refresh set} of tokens.
Crucially, we find that the intuitive heuristic of refreshing visually drifted tokens plateaus far below the dense baseline, even with an oracle predicting ground-truth KV drift.
Downstream action accuracy is instead governed by where the action expert attends, not by what moved.
WAM-Cache therefore selects the refresh set by uniting the action expert's cross-attention with visual latent surprise, complemented by a strict age bound that suppresses compounding error.
On Fast-WAM, WAM-Cache cuts video DiT prefill FLOPs by 32--42\% across RoboTwin~2.0, LIBERO, and real-world experiments, while staying within 0.7--1.8 percentage points of the dense policy in simulation and 2.5 points on a real robot.
\end{abstract}

\section{Introduction}
\label{sec:introduction}

World Action Models (WAMs) have emerged as a promising paradigm for generalist robot manipulation~\citep{wang2026worldactionmodelsnextfrontier,ye2026worldactionmodelszeroshot,fastwam2026,lingbot-va2026}.
Unlike vision-language-action (VLA) policies that map observations to actions through a vision-language backbone~\citep{black2024pi_0,zitkovich2023rt,kim2024openvla}, a WAM pairs a Diffusion Transformer (DiT)~\citep{peebles2023scalable} pretrained for video generation~\citep{wan2025,nvidia2025cosmos} with an action expert that conditions on its representations.
This design grounds control in physical dynamics learned from video, substantially enhancing generalization across unseen scenes and objects~\citep{du2023learninguniversalpoliciestextguided,wu2023unleashing,hu2024video,zhu2025unifiedworldmodelscoupling,ye2026worldactionmodelszeroshot}.
Yet, this capability comes at a steep computational cost.
When operating in closed loop, a WAM repeatedly observes the environment and replans action chunks, requiring the video DiT to run at every control cycle.
This backbone accounts for the vast majority of a WAM's parameters and compute.

\begin{figure}[t]
    \centering
    \includegraphics[width=\textwidth]{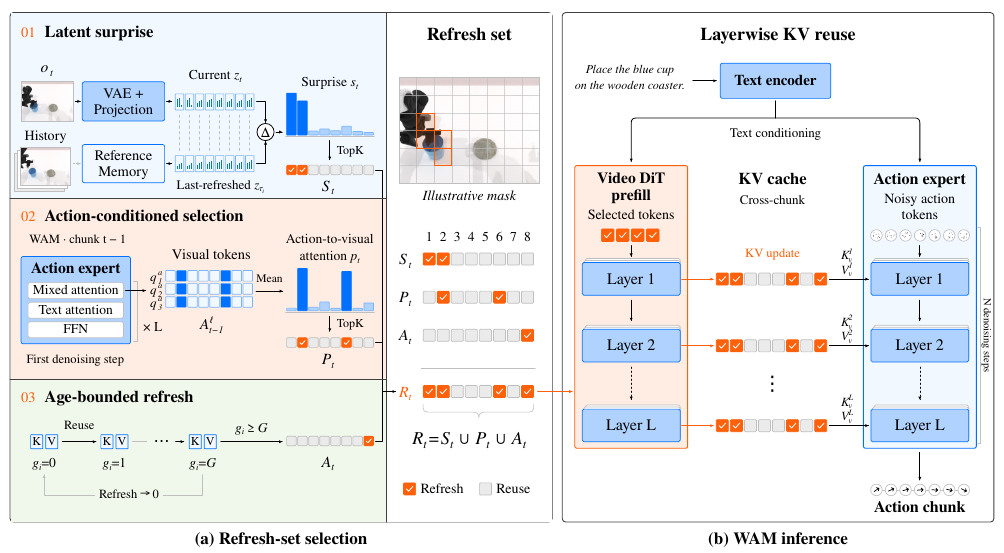}
    \caption{\textbf{The WAM-Cache inference pipeline.} (a)~\textbf{Sparse refresh selection at chunk $t$}: three complementary signals construct the refresh set $\mathcal{R}_t$. Latent surprise measures feature drift from a token's reference embedding; action-conditioned selection identifies tokens heavily attended to by the action expert in the preceding chunk; and an age counter forces recomputation after $G$ consecutive reuses. (b)~\textbf{Sparse prefill and mixed-cache execution}: the video DiT executes a sparse prefill exclusively over $\mathcal{R}_t$, updating layerwise keys and values in-place. The action expert then denoises the next action chunk over the resulting mixed cache, querying fresh representations for $\mathcal{R}_t$ alongside retained keys and values for unselected tokens.}
    \label{fig:wam-cache-method}
\end{figure}

Prior efforts to mitigate this cost generally fall into two regimes: retraining-based and training-free approaches.
Retraining-based methods distill the model~\citep{akbari2026flashwam}, compress the backbone~\citep{li2026efficientwam,zhang2026imagewam}, or simplify action decoding~\citep{li2026lightwam,ma2026fasterwamdot}.
Training-free approaches, by contrast, focus primarily on caching intermediate denoiser representations across steps or chunks~\citep{ma2023deepcache,zou2024toca,zhao2026c3ache,zeng2026xcache}.
Reducing the video DiT's forward pass, the largest cost term, has so far required retraining a backbone with billions of parameters.
To our knowledge, existing training-free methods leave this bottleneck entirely untouched.
This forward pass—the \emph{video DiT prefill}—encodes the current observation into layerwise key--value (KV) pairs, which the action expert queries while denoising an action chunk.
As action experts adopt fewer denoising steps to cut latency~\citep{wang2024onedp,dong2026flowtoonestep,chen2026onestepvisuomotor}, the prefill increasingly dominates the per-chunk computational cost.
On Fast-WAM, for example, the prefill accounts for 50\% of DiT FLOPs at the default ten action-denoising steps and surges to 83.5\% at two steps, with virtually no loss in task success (Appendix~\ref{app:steps}).

In large language model (LLM) inference, such prefill redundancy is routinely eliminated by KV caching~\citep{pope2023efficiently,zhang2023h2o,li2024snapkv,mao2026triattention}: prompt representations are computed once, and subsequent decoding steps simply attend to the cached states.
That reuse is exact because the prompt never changes, whereas a WAM ingests a fresh observation at every chunk.
However, most of that observation remains static.
Manipulation is local: from one chunk to the next, motion is largely restricted to the end-effector and target object, rendering most of the visual field redundant.
Despite this spatial locality, the prefill recomputes every visual token at every layer.
Crucially, the action expert interacts with the video DiT solely through its key--value states, making KV reuse architecturally transparent.
The core challenge is therefore to decide which tokens can stay cached across chunks and which must be recomputed.

To this end, we introduce \textbf{WAM-Cache} (Figure~\ref{fig:wam-cache-method}), a training-free caching framework that retains layerwise key--value representations across inference chunks, recomputing only a sparse \emph{refresh set} of visual tokens at each chunk.
The entire design hinges on choosing this refresh set, yet the intuitive rule of refreshing what has changed fails in this setting.
Refreshing by change alone causes a steep drop in RoboTwin~2.0~\citep{chen2025robotwin} success relative to the dense model (Figure~\ref{fig:wam-cache-components}).
Even an oracle with access to the ground-truth key--value drift—refreshing the tokens that drifted most—improves on this rule only marginally (Section~\ref{sec:ablation}).
This performance ceiling reveals that the downstream impact of a stale KV pair is dictated by how the action expert consumes the cache.
To a first approximation, the action depends on each cached pair in proportion to the attention allocated to it: attention gauges downstream sensitivity to stale features, whereas drift measures only local representation error.
WAM-Cache therefore guides token refresh with two content signals: one tracking feature drift, and the other tracking downstream policy sensitivity.
\emph{Latent surprise} approximates drift by measuring how far a token's input embedding has shifted since its last refresh.
\emph{Action-conditioned selection} refreshes the tokens most heavily attended to by the action expert in the preceding chunk—leveraging cross-attention maps that are already computed, at zero extra cost.
These two signals select largely disjoint token subsets, which WAM-Cache combines via a simple union.
However, because both signals rely on rank-based selection, an unselected token can repeatedly miss the cutoff while its cached error steadily compounds, thereby preserving a clear gap to the dense model.
An \emph{age bound} therefore forces a refresh for any token reused across a fixed number of consecutive chunks.
This caps the staleness of every pair regardless of rank, nearly closing that gap (Figure~\ref{fig:wam-cache-components}).
Consequently, the prefill runs exclusively on this sparse refresh set, allowing the action expert to operate over the mixed cache without any modification.
As a result, WAM-Cache requires no retraining and composes seamlessly with existing orthogonal speedups, including diffusion step reduction, model distillation, and denoiser caching.

On Fast-WAM, WAM-Cache cuts video DiT prefill FLOPs by 41.6\% on RoboTwin~2.0 at a cost of 1.8 points in success, whereas token-reduction baselines at a matched budget lose 4 to 38 points.
The prefill's CUDA latency drops by $1.23\times$ at the native token count and by $1.67\times$ at a $9\times$ larger visual context.
WAM-Cache also removes 32.3\% of the prefill FLOPs on LIBERO~\citep{liu2023libero} within 0.7 points of the dense model.
On two real-world manipulation tasks with an AIRBOT Play arm, it removes 40.2\% of the prefill FLOPs while achieving success comparable to the dense model, 82.5\% against 85.0\%.

Our contributions are summarized as follows:
\begin{itemize}
    \item We propose \textbf{WAM-Cache}, to the best of our knowledge, the first training-free key--value reuse framework for the video DiT prefill of World Action Models (Section~\ref{sec:method}).

    \item We show that cache refresh must follow policy attention rather than visual drift: change-based rules plateau 25 points below the dense model even with an oracle, while an age bound strictly caps staleness to nearly close the gap (Section~\ref{sec:ablation}).

    \item Across RoboTwin~2.0, LIBERO, and real-world robot experiments, WAM-Cache reduces prefill FLOPs by 32--42\% while staying close to the dense model (Section~\ref{sec:experiments}).
\end{itemize}

\section{Related Work}
\label{sec:related-work}

\textbf{World Action Models.}
Generative video models have been used in robotics for trajectory planning~\citep{du2023learninguniversalpoliciestextguided,zhou2024robodreamer,bharadhwaj2024gen2act,zhen20243d}, policy pretraining~\citep{wu2023unleashing,cheang2024gr2generativevideolanguageactionmodel,hu2024video}, and data generation or policy evaluation~\citep{jang2025dreamgenunlockinggeneralizationrobot,team2026gigaworld1}, and now serve as policy backbones~\citep{zhu2025unifiedworldmodelscoupling,li2025unified,liang2025videogenerators,pai2025mimicvideo,kim2026cosmos,ge2025,cen2025WorldVLA}.
Recent WAMs~\citep{lyu2025dywa,chen2026lawam} mostly pair video DiTs with action decoders~\citep{lingbot-va2026,zhang2026lingbotva2,bi2025motusunifiedlatentaction,zhou2026zerowamincontextworldactionmodeling,team2026gigaworldpolicy,wang2026openwam,team2026motubrain,huang2026longwam}.
Most of them generate future video before or alongside the actions at inference.
Fast-WAM~\citep{fastwam2026} and GigaWorld-Policy~\citep{gigaworldpolicy2026} show that this future imagination can be dropped with little loss in task success, which cuts inference to a single video DiT prefill that encodes the current observation into layerwise key--value pairs for the action expert.
This prefill is now the dominant per-chunk cost, and to the best of our knowledge, WAM-Cache is the first method to make it sparse.

\textbf{Efficient Policy Inference and Token Reuse.}
Retraining approaches distill~\citep{akbari2026flashwam} or compress~\citep{li2026efficientwam,li2026lightwam,ma2026fasterwamdot} the model, reduce denoising steps~\citep{wang2024onedp,dong2026flowtoonestep,chen2026onestepvisuomotor,zhang2026lingbotva2}, or make future imagination cheaper or adaptive~\citep{zhang2026imagewam,li2026lawa,zhao2026fasterwamfc,tang2026adawam,yan2026flexpi,sun2026sants}, and asynchronous schemes overlap inference with execution~\citep{guo2026reflex,li2026flashvla,li2026fbfm}.
Training-free caching reuses denoiser features across denoising steps~\citep{ma2023deepcache,zou2024toca,feng2026worldcache} or inference chunks~\citep{zeng2026xcache,zhao2026c3ache}, but the WAM video prefill is a single pass with no denoising trajectory to reuse.
Token reduction prunes~\citep{chen2024fastv} or merges~\citep{bolya2022tome} tokens within a frame, while Eventful Transformer~\citep{dutson2023eventful} and VLA-Cache~\citep{xu2025vlacache,wu2026temporalredundancyvla,wu2026gatedvlacache,luo2026gatenotcache} reuse static tokens across frames, selecting them by change or by attention inside the unified backbone, with no bound on how long a token is reused.
In a WAM, however, visual prefill and action decoding are handled by decoupled modules. WAM-Cache therefore guides token refresh by downstream policy attention rather than visual drift, while enforcing an age bound that strictly caps staleness.

\section{Preliminaries}
\label{sec:prelim}

\textbf{World Action Models.}
A world action model (WAM) pairs a video DiT with an action-expert DiT for language-conditioned robot control.
The model maps an observation $o_t$ and a text instruction $c$ to an action chunk, and the robot executes a prefix of the chunk before observing again.
We call one such cycle an \emph{inference chunk} and index it by $t$.
A VAE encodes $o_t$ into a latent, and a patch-embedding projection maps the latent to $S_v$ visual tokens of width $d$, $\mathbf{z}_t\in\mathbb{R}^{S_v\times d}$.
We write $\mathbf{z}_t=\operatorname{Enc}(o_t)$ for this composition.
Throughout, a subscript $t$ marks a quantity computed at chunk $t$, and $[i]$ indexes token $i$.
The video DiT processes $\mathbf{z}_t$ in a single forward pass.
At each of its $N$ denoising steps the action expert cross-attends to the prefill's keys and values to predict the action chunk.

\textbf{Key--Value Caching in DiTs.}
A Diffusion Transformer (DiT)~\citep{peebles2023scalable} stacks $L$ modules over a sequence of tokens.
Module $l$ applies a self-attention layer $\mathcal{F}^l_{\mathrm{SA}}$, a cross-attention layer $\mathcal{F}^l_{\mathrm{CA}}$ that attends to keys and values projected from the text embeddings, and a feed-forward layer $\mathcal{F}^l_{\mathrm{MLP}}$, each with a residual connection.
Self-attention at layer $l$ of the video DiT projects its input $\mathbf{H}^l_t$ into queries $\mathbf{Q}^l_t=\mathbf{H}^l_t\mathbf{W}^l_Q$, keys $\mathbf{K}^l_t=\mathbf{H}^l_t\mathbf{W}^l_K$, and values $\mathbf{V}^l_t=\mathbf{H}^l_t\mathbf{W}^l_V$, and each query attends over all $S_v$ keys and values.
A dense prefill thus produces $S_v$ key--value pairs at each of the $L$ layers.
Key--value caching stores these pairs so that a later pass can reuse them instead of recomputing them.

\section{Method}
\label{sec:method}
WAM-Cache accelerates inference by retaining layerwise key--value representations across chunks, recomputing only a sparse \emph{refresh set} of visual tokens at each chunk (illustrated in Figure~\ref{fig:wam-cache-method}).
In what follows, we first formalize the cache state, the per-chunk execution procedure, and the resulting computational savings (Section~\ref{sec:method-overview}), and then detail the three complementary signals that construct the refresh set (Sections~\ref{sec:method-surprise}--\ref{sec:method-age}).

\subsection{Overview: Staleness-Bounded KV Reuse}
\label{sec:method-overview}
\textbf{Per-token cache state.}
Across consecutive inference chunks, WAM-Cache maintains the layerwise key--value representations alongside lightweight tracking metadata for each visual token $i \in \{1, \ldots, S_v\}$:
(i) the cached key--value pairs $\{(\mathbf{K}^l[i], \mathbf{V}^l[i])\}_{l=1}^L$,
(ii) a reference embedding $\mathbf{z}_{\tau_i}[i] \in \mathbb{R}^d$ stored at its most recent refresh chunk $\tau_i$,
(iii) an age counter $g_i \in \mathbb{N}$ tracking consecutive chunks of reuse, and
(iv) the action expert's attention mass $\mathbf{p}_t[i]$ from the preceding chunk (Section~\ref{sec:method-attention}).
At episode onset ($t=0$), WAM-Cache executes a standard dense prefill, initializing $\tau_i = 0$ and $g_i = 0$ for all tokens.
In subsequent chunks, these scalar tracking variables are updated in place without logging historical states.

\textbf{Per-chunk procedure.}
For each subsequent chunk $t > 0$, WAM-Cache extracts patch embeddings $\mathbf{z}_t = \operatorname{Enc}(o_t)$ and constructs a sparse \emph{refresh set}
\begin{equation}
    \mathcal{R}_t \;=\; \mathcal{S}_t \cup \mathcal{P}_t \cup \mathcal{A}_t \;\subseteq\;\{1,\ldots,S_v\},
    \label{eq:wam-refresh-set}
\end{equation}
formed by the union of a latent-surprise set $\mathcal{S}_t$ (Section~\ref{sec:method-surprise}), an action-attention set $\mathcal{P}_t$ (Section~\ref{sec:method-attention}), and an age set $\mathcal{A}_t$ enforcing periodic renewal (Section~\ref{sec:method-age}).
The video DiT prefill then executes exclusively over the tokens in $\mathcal{R}_t$.
At each block $l$, the hidden states $\mathbf{H}^l_t[\mathcal{R}_t]$ are recomputed, and their linear projections overwrite the retained cache in place, while unselected tokens maintain their previously stored representations:
\begin{equation}
    \mathbf{K}^l_t[i]=\begin{cases}
        \mathbf{H}^l_t[i]\,\mathbf{W}^l_K, & i\in\mathcal{R}_t,\\
        \mathbf{K}^l_{\tau_i}[i], & i\notin\mathcal{R}_t,
    \end{cases}
    \qquad\text{and likewise for }\mathbf{V}^l_t.
    \label{eq:wam-partial-kv}
\end{equation}
Meanwhile, queries are formed exclusively for refreshed tokens ($i \in \mathcal{R}_t$) and attend over all $S_v$ keys and values, allowing newly recomputed features to integrate both updated dynamics and static context.
Following the pass, WAM-Cache updates the refresh timestamps $\tau_i \leftarrow t$, increments the age counters via Eq.~(\ref{eq:wam-age-update}), and records the action expert's cross-attention for chunk $t+1$.
Because no token can be reused for more than $G$ consecutive chunks, this mechanism guarantees \emph{staleness-bounded} reuse.
Algorithm~\ref{alg:wam-cache} in Appendix~\ref{app:algorithm} summarizes the per-chunk procedure.

\textbf{Prefill cost.}
\label{sec:method-cost}
With $S_c$ text tokens and feed-forward width $d_f$, a dense prefill of one video DiT module over $S_v$ visual tokens costs, in FLOPs,
\begin{equation}
    C_{\mathrm{dense}}
    \approx\underbrace{8S_v d^2+4S_v^2 d}_{\mathcal{F}_{\mathrm{SA}}}
    +\underbrace{4S_v d^2+4S_v S_c d}_{\mathcal{F}_{\mathrm{CA}}}
    +\underbrace{4S_v d\,d_f}_{\mathcal{F}_{\mathrm{MLP}}}
    \label{eq:wam-dense-cost}
\end{equation}
Let $\rho_t=|\mathcal{R}_t|/S_v$ be the recompute ratio of chunk $t$, and let $\bar\rho$ be its average over the chunks of an episode after the first.
Since only the tokens in $\mathcal{R}_t$ form queries and every term of Eq.~(\ref{eq:wam-dense-cost}) is linear in the number of query tokens, the per-module cost and the saving are
\begin{equation}
    C(\rho_t)=\rho_t\,C_{\mathrm{dense}},
    \qquad
    \Delta C(\rho_t)=(1-\rho_t)\,C_{\mathrm{dense}}.
    \label{eq:wam-cost}
\end{equation}
The saving is therefore set entirely by how $\mathcal{R}_t$ is chosen.
Appendix~\ref{app:cost} derives both equations, shows that the selection signals add negligible overhead, and bounds $\bar\rho$.

\begin{wrapfigure}{r}{0.5\textwidth}
    \vspace{-6pt}
    \centering
    \includegraphics[width=\linewidth]{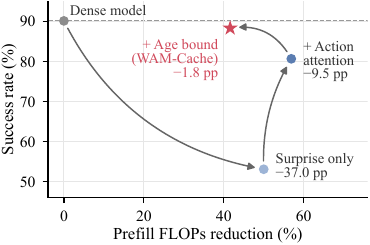}
    \vspace{-20pt}
    \caption{Sequential ablation of caching signals on clean RoboTwin~2.0. The first two configurations run without the age bound.}
    \label{fig:wam-cache-components}
    \vspace{-22pt}
\end{wrapfigure}
The next three subsections build the refresh rule step by step.
We start from the intuitive rule of refreshing what changed, and each later signal addresses a failure of the preceding configuration.
Figure~\ref{fig:wam-cache-components} reports this cumulative construction.

\subsection{Latent Surprise}
\label{sec:method-surprise}
We begin with the most direct refresh criterion: recompute a token when its input has changed.
Target objects shift, or the robot arm traverses previously static visual regions.
Even a sequence of minor displacements can separate the current observation from its cached representation.

\textbf{Surprise.}
For each token $i$, we compare its current embedding with its reference $\mathbf{z}_{\tau_i}[i]$ and score the normalized distance
\begin{equation}
\mathbf{s}_t[i] \;=\; \frac{\bigl\lVert \mathbf{z}_t[i] - \mathbf{z}_{\tau_i}[i] \bigr\rVert_2}
{\tfrac{1}{2}\bigl( \lVert \mathbf{z}_t[i] \rVert_2 + \lVert \mathbf{z}_{\tau_i}[i] \rVert_2 \bigr) + \epsilon},
\label{eq:surprise}
\end{equation}
where $\epsilon$ prevents division by zero.
The latent-surprise set $\mathcal{S}_t=\operatorname{TopK}(\mathbf{s}_t,b_s)$ keeps the highest-scoring tokens under a budget $b_s$, where $\operatorname{TopK}(\mathbf{s},b)$ returns the indices of the $\max\{1,\operatorname{round}(bS_v)\}$ largest entries of $\mathbf{s}$.
The reference is updated only when its token is recomputed, so the score measures displacement from the observation that the retained pair represents, not from the preceding frame.

Surprise alone is insufficient.
On clean RoboTwin~2.0, this rule reaches 53.1\% success without the age bound, 37.0 points below the dense model (Figure~\ref{fig:wam-cache-components}).
We attribute this performance gap to regions that change little in latent space yet matter for contact or alignment.
Latent displacement does not capture how heavily the action expert relies on a token, so a change-only ranking assigns these regions low priority.

\subsection{Action-Conditioned Selection}
\label{sec:method-attention}
The first failure concerns regions that stay visually stable while the policy depends on them.
Near a grasp, for example, the target object barely moves, yet its exact position determines the next action.
The action expert's attention indicates which visual tokens its preceding prediction relied on.

\textbf{Action attention.}
Let $\mathbf{A}^{l}_{t-1} \in \mathbb{R}^{S_a \times S_v}$ be the head-averaged cross-attention from the $S_a$ action queries to the visual tokens at action-expert layer $l\in\{1,\ldots,L_a\}$, recorded at the first action-denoising step of chunk $t-1$.
The attention mass on token $i$ is averaged over layers and queries,
\begin{equation}
\mathbf{p}_t[i] \;=\; \frac{1}{L_a S_a} \sum_{l=1}^{L_a} \sum_{m=1}^{S_a}
\mathbf{A}^{l}_{t-1}[m, i],
\label{eq:action-mass}
\end{equation}
and the action-attention set $\mathcal{P}_t=\operatorname{TopK}(\mathbf{p}_t,b_a)$ keeps the highest-scoring tokens under a budget $b_a$.
The score depends only on the previous chunk, so it is available before the current prefill at no extra cost.

The union $\mathcal{S}_t\cup\mathcal{P}_t$ thus simultaneously captures dynamic physical changes and policy-critical static regions, although inherited attention can lag a transition between manipulation phases.
The joint selection raises success to 80.7\%, 27.6 points above surprise alone but still 9.5 points below the dense model.
We therefore look for a third component that depends on neither content score.

\subsection{Age-Bounded Refresh}
\label{sec:method-age}
The second failure arises from repeated reuse.
Both content signals are self-referential: each decides which tokens to refresh from state that earlier refresh decisions produced.
A refreshed query attends over retained keys and values from unrefreshed tokens, and the subsequent attention score is computed from a partially refreshed context.
Because both signals rely on rank-based selection, an unselected token can repeatedly miss the cutoff while its cached error steadily compounds.
Neither a low surprise score nor a weak attention weight justifies retaining a token indefinitely.

\textbf{Age bound.}
The age set collects every token whose counter has reached the bound $G$,
\begin{equation}
    \mathcal{A}_t \;=\; \{\, i : g_i \ge G \,\},
    \label{eq:wam-age-set}
\end{equation}
and these tokens are refreshed regardless of their scores.
We write $G=0$ for the rule disabled, in which case $\mathcal{A}_t=\varnothing$.
After the pass, the counters update as
\begin{equation}
    g_i \;\leftarrow\;
    \begin{cases}
        0,       & i \in \mathcal{R}_t, \\
        g_i + 1, & i \notin \mathcal{R}_t.
    \end{cases}
    \label{eq:wam-age-update}
\end{equation}
The age bound operates only on state already stored in the cache and adds no forward computation.
Yet it nearly closes the gap to the dense model, leaving 1.8 points, while still removing 41.6\% of the prefill FLOPs.
These forced refreshes target only tokens that neither content score has selected for $G$ chunks and raise $\bar\rho$ by at most $1/(G+1)$ (Appendix~\ref{app:cost}).
However, the age bound limits how long a pair can stay stale, not how large its error can grow.

\section{Experiments}
\label{sec:experiments}
To demonstrate the effectiveness of WAM-Cache, we evaluate our method in both simulation and real-world environments.
In the simulation environment, we evaluate WAM-Cache on Fast-WAM~\citep{fastwam2026}, using the LIBERO benchmark~\citep{liu2023libero} and the RoboTwin~2.0 benchmark~\citep{chen2025robotwin}.
In the real-world environment, we deploy Fast-WAM with WAM-Cache on an AIRBOT Play arm for two manipulation tasks.

\subsection{Implementation Details}
\label{sec:impl-details}

\label{sec:impl-arch}
We apply WAM-Cache to Fast-WAM~\citep{fastwam2026}, which pairs the Wan2.2-5B video DiT~\citep{wan2025} with a 1B-parameter action-expert DiT and encodes only the current observation at test time (architecture details in Appendix~\ref{app:base-model}).
All models run in bfloat16 on a single NVIDIA RTX PRO 6000 Blackwell GPU.

\label{sec:impl-inference}
\textbf{Action-denoising steps.}
We use $N=2$ action-denoising steps by default and follow Fast-WAM in all other inference settings.
For the dense model, two steps stay within 1.5 points of ten on RoboTwin~2.0, whereas one step loses more than 20, and at both step counts the video DiT prefill is the largest per-chunk cost.
We also evaluate WAM-Cache separately at Fast-WAM's original $N=10$, where it stays within 1.1 points of the dense model (Appendix~\ref{app:steps}).

\textbf{Hyperparameters.}
We fix $G=2$ and report two budget settings per benchmark: a default one, to whose prefill FLOPs the baselines are calibrated, and a larger one that approaches the dense model.
The exact values of $(b_a,b_s)$ are listed with each table, and Figure~\ref{fig:wam-cache-sweeps} shows how they are chosen.

\subsection{Simulation Experiments}
\label{sec:sim-exp}

\label{sec:sim-setup}
\textbf{Benchmarks.}
RoboTwin~2.0 contains 50 bimanual tasks, each evaluated under clean and randomized scenes, and LIBERO contains four single-arm suites (Spatial, Object, Goal, Long) of ten tasks each.
We use the released Fast-WAM checkpoints and the Fast-WAM evaluation protocol.
Episode counts are listed with each table, and execution details are given in Appendix~\ref{app:base-model}.

\textbf{Baselines.}
We compare against four training-free token-reduction methods, adapted to the same video DiT prefill at a matched FLOPs budget.
VLA-Cache~\citep{xu2025vlacache} and Eventful Transformer~\citep{dutson2023eventful} reuse tokens across frames, while FastV~\citep{chen2024fastv} and ToMe~\citep{bolya2022tome} prune or merge tokens within a frame.
VLA-Cache is designed for VLA backbones, so we adapt it to the WAM video DiT prefill.
Tokens whose image patches stay static between replans reuse their cached keys and values, except those with high video-to-text attention.
Its threshold is calibrated to match the recompute ratio $\bar\rho$ of WAM-Cache (Appendix~\ref{app:base-model}).

\textbf{Metrics.}
We report success rate (SR), prefill FLOPs averaged over all chunks after the first, and CUDA latency of the same stage under CUDA-graph replay, following prior VLM and VLA acceleration work~\citep{chen2024fastv,xu2025vlacache}.

\label{sec:sim-main}
\textbf{Main results.}
Table~\ref{tab:wam-cache-main} reports RoboTwin~2.0.
At a matched FLOPs budget, WAM-Cache removes 41.6\% and 39.3\% of the prefill FLOPs under clean and randomized scenes and stays within 1.8 and 1.3 points of the dense prefill, whereas every baseline loses at least 4.0 points.
VLA-Cache, the strongest baseline, also combines a change signal with attention, but its attention comes from the language tokens rather than the action expert, and its reuse is unbounded.
With a larger budget, WAM-Cache comes within 0.1 points of the dense prefill under both conditions while still removing 28.2\% and 37.0\% of the prefill FLOPs.
Table~\ref{tab:wam-cache-libero} shows that the same rules transfer to LIBERO with only $b_a$ raised: at a 32.3\% reduction WAM-Cache averages 95.85\% against 96.50\% for the dense prefill and outperforms every baseline at a comparable budget, and at a 26.1\% reduction it matches the dense average.
On LIBERO the wrist view fills half of the frame, so the policy attends to more tokens per chunk.
Most of the remaining loss falls on Long, whose multi-phase tasks expose the lag of inherited attention at phase transitions (Section~\ref{sec:method-attention}).

\begin{table}[t]
    \centering
    \caption{Results on RoboTwin~2.0 (50 tasks, 100 episodes per task). FLOPs is the per-chunk video DiT prefill cost, and $\Delta$FLOPs and $\Delta$SR are relative to the dense prefill. Each baseline is calibrated to the prefill FLOPs of the first WAM-Cache row of each condition. The two WAM-Cache rows use $(b_a,b_s)=(0.4,0.1)$ and $(0.6,0.1)$ under clean observations and $(0.4,0.2)$ and $(0.4,0.3)$ under randomization.}
    \label{tab:wam-cache-main}
    \small
    \setlength{\tabcolsep}{3pt}
    \begin{tabular}{l S[table-format=1.3] S[table-format=-2.1] S[table-format=2.2] S[table-format=-2.1] S[table-format=1.3] S[table-format=-2.1] S[table-format=2.2] S[table-format=-2.1]}
        \toprule
        & \multicolumn{4}{c}{Clean} & \multicolumn{4}{c}{Randomized} \\
        \cmidrule(lr){2-5} \cmidrule(lr){6-9}
        \multirow{2}{*}{Method} & {FLOPs} & {$\Delta$FLOPs} & {SR} & {$\Delta$SR} & {FLOPs} & {$\Delta$FLOPs} & {SR} & {$\Delta$SR} \\
        & {(T) $\downarrow$} & {(\%)} & {(\%) $\uparrow$} & {(pp)} & {(T) $\downarrow$} & {(\%)} & {(\%) $\uparrow$} & {(pp)} \\
        \midrule
        Fast-WAM (dense prefill) & 1.053 & 0.0 & 90.16 & 0.0 & 1.053 & 0.0 & 89.66 & 0.0 \\
        \midrule
        + FastV~\citep{chen2024fastv} & 0.618 & -41.3 & 62.72 & -27.4 & 0.642 & -39.0 & 60.18 & -29.5 \\
        + ToMe~\citep{bolya2022tome} & 0.622 & -41.0 & 77.92 & -12.2 & 0.639 & -39.3 & 74.14 & -15.5 \\
        + Eventful~\citep{dutson2023eventful} & 0.623 & -40.8 & 52.94 & -37.2 & 0.641 & -39.2 & 51.36 & -38.3 \\
        + VLA-Cache~\citep{xu2025vlacache} & 0.623 & -40.8 & 86.16 & -4.0 & 0.649 & -38.4 & 84.26 & -5.4 \\
        \midrule
        \textbf{+ WAM-Cache (ours)} & \bfseries 0.615 & \bfseries -41.6 & \bfseries 88.36 & \bfseries -1.8 & \bfseries 0.639 & \bfseries -39.3 & \bfseries 88.32 & \bfseries -1.3 \\
        \textbf{+ WAM-Cache (ours), larger budget} & \bfseries 0.756 & \bfseries -28.2 & \bfseries 90.08 & \bfseries -0.1 & \bfseries 0.663 & \bfseries -37.0 & \bfseries 89.56 & \bfseries -0.1 \\
        \bottomrule
    \end{tabular}
\end{table}

\begin{table}[t]
    \centering
    \caption{Results on LIBERO (50 episodes per task, 500 per suite). Each baseline is calibrated to the prefill FLOPs of the first WAM-Cache row. The two WAM-Cache rows use $(b_a,b_s)=(0.5,0.1)$ and $(0.6,0.1)$.}
    \label{tab:wam-cache-libero}
    \small
    \setlength{\tabcolsep}{3.5pt}
    \begin{tabular}{l S[table-format=2.2] S[table-format=2.2] S[table-format=2.2] S[table-format=2.2] S[table-format=2.2] S[table-format=1.3] S[table-format=-2.1]}
        \toprule
        & \multicolumn{5}{c}{SR (\%) $\uparrow$} & {FLOPs} & {$\Delta$FLOPs} \\
        \cmidrule(lr){2-6}
        Method & {Spatial} & {Object} & {Goal} & {Long} & {Avg.} & {(T) $\downarrow$} & {(\%)} \\
        \midrule
        Fast-WAM (dense prefill) & 96.80 & 99.40 & 96.00 & 93.80 & 96.50 & 0.859 & 0.0 \\
        \midrule
        + FastV~\citep{chen2024fastv} & 73.20 & 93.60 & 75.40 & 73.00 & 78.80 & 0.588 & -31.5 \\
        + ToMe~\citep{bolya2022tome} & 94.20 & 97.60 & 92.20 & 83.80 & 91.95 & 0.587 & -31.7 \\
        + Eventful~\citep{dutson2023eventful} & 95.40 & 97.60 & 94.20 & 87.40 & 93.65 & 0.599 & -30.2 \\
        + VLA-Cache~\citep{xu2025vlacache} & 96.00 & 98.40 & 94.80 & 90.20 & 94.85 & 0.588 & -31.5 \\
        \midrule
        \textbf{+ WAM-Cache (ours)} & \bfseries 96.20 & \bfseries 99.20 & \bfseries 97.60 & \bfseries 90.40 & \bfseries 95.85 & \bfseries 0.582 & \bfseries -32.3 \\
        \textbf{+ WAM-Cache (ours), larger budget} & \bfseries 96.80 & \bfseries 99.60 & \bfseries 95.60 & \bfseries 94.00 & \bfseries 96.50 & \bfseries 0.635 & \bfseries -26.1 \\
        \bottomrule
    \end{tabular}
\end{table}

\begin{figure}[t]
    \centering
    \begin{minipage}[c]{0.5\textwidth}
        \centering
        \includegraphics[width=\linewidth]{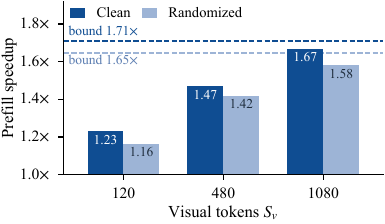}
    \end{minipage}\hfill
    \begin{minipage}[c]{0.46\textwidth}
        \centering
        \small
        \setlength{\tabcolsep}{4pt}
        \begin{tabular}{l r r}
            \toprule
            Selection rule & SR (\%) $\uparrow$ & $\bar\rho$ $\downarrow$ \\
            \midrule
            Random & 26.88 & 0.571 \\
            Pixel difference & 60.40 & 0.625 \\
            Oracle K/V drift & 65.34 & 0.599 \\
            Surprise only & 60.62 & 0.615 \\
            Action attention only & 81.46 & 0.643 \\
            \textbf{Attention $\cup$ surprise (ours)} & \textbf{88.36} & \textbf{0.584} \\
            \bottomrule
        \end{tabular}
    \end{minipage}\\[-10pt]
    \begin{minipage}[t]{0.5\textwidth}
        \captionof{figure}{Prefill speedup in CUDA latency of the default clean and randomized configurations. Dashed lines mark the bound $1/\bar\rho$.}
        \label{fig:wam-cache-latency}
    \end{minipage}\hfill
    \begin{minipage}[t]{0.46\textwidth}
        \captionof{table}{Ablation of refresh rules on clean RoboTwin~2.0 (content budget 0.5, $G=2$).}
        \label{tab:wam-cache-ablation}
    \end{minipage}
\end{figure}

\textbf{CUDA latency.}
\label{sec:sim-latency}
Figure~\ref{fig:wam-cache-latency} reports the prefill speedup of the first clean and randomized configurations of Table~\ref{tab:wam-cache-main} at the native $S_v=120$ and at $4\times$ and $9\times$ larger visual contexts.
No checkpoint exists at the larger sizes, so we replay the refresh ratios measured at $S_v=120$ and measure cost only.
By Eq.~(\ref{eq:wam-cost}) the speedup is bounded by $1/\bar\rho$.
Under clean observations WAM-Cache speeds up the prefill by $1.23\times$, $1.47\times$, and $1.67\times$ at $S_v=120$, 480, and 1080, reaching 98\% of the $1.71\times$ bound at the largest size, and the randomized configuration reaches $1.58\times$, 96\% of its bound.
At $S_v=120$ only about 70 tokens are refreshed per chunk, so weight loading and graph replay take a fixed share of each layer's time.
The gap to the bound therefore closes as the visual context grows, which is the regime in which the prefill dominates.
As WAMs move toward higher resolutions, more camera views, and longer observation histories, the realized speedup of WAM-Cache should increasingly approach its FLOPs bound.

\subsection{Ablation Study}
\label{sec:ablation}
Table~\ref{tab:wam-cache-ablation} compares six refresh rules on clean RoboTwin~2.0 under the same content budget, so the rules differ only in \emph{which} tokens they refresh.
The single-signal rules spend the whole budget on one score, $b_s=0.5$ or $b_a=0.5$, and WAM-Cache splits it as $(b_a,b_s)=(0.4,0.1)$.
The K/V-drift oracle refreshes the tokens whose cached keys and values drift most from a dense prefill of the new frame.
It is an upper bound for any rule that predicts cache drift, not a practical method.

\textbf{Which tokens are refreshed matters more than how many.}
At a matched recompute ratio, random selection reaches only 26.9\%.
Every informed rule does far better.

\textbf{Change-based rules plateau.}
Three rules select tokens by change: pixel difference (the change signal of VLA-Cache), latent surprise, and the K/V-drift oracle.
All three stay between 60.4\% and 65.3\%.
Even ground-truth knowledge of cache drift leaves a 25-point gap to the dense model.
Drift is therefore a weak proxy for how much a stale token harms the action (Section~\ref{sec:method-surprise}).

\textbf{Action attention recovers most of the gap.}
Action attention alone reaches 81.5\%.
Shifting 0.1 of the budget to surprise raises success to 88.4\%, since the two signals select largely disjoint tokens.
The union is also the cheapest informed rule, with $\bar\rho=0.584$ against 0.643 and 0.615.
The surprise set changes from chunk to chunk, so it resets many counters before the age bound forces a refresh.
Tokens selected by both signals are also counted only once.

\textbf{Age bounding breaks self-referential error accumulation.}
Without the age bound, content signals select by rank, so low-priority tokens can escape refresh indefinitely.
Their error compounds and leaves a 9.5-point gap to the dense model.
Setting $G=2$ bounds the staleness of every token.
It recovers 7.7 points of success and brings the policy within 1.8 points of the dense prefill (Figure~\ref{fig:wam-cache-components}).

\begin{figure}[t]
    \centering
    \includegraphics[width=\textwidth]{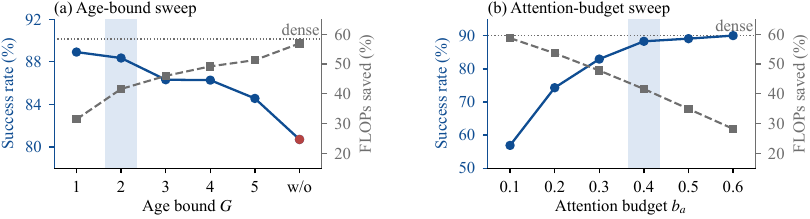}
    \vspace{-16pt}
    \caption{Hyperparameter sweeps on clean RoboTwin~2.0. (a) Age bound $G$ at $(b_a,b_s)=(0.4,0.1)$, where \emph{w/o} runs without the bound. (b) Attention budget $b_a$ at $b_s=0.1$ and $G=2$. Blue circles (left axis) are the success rate and gray squares (right axis) the prefill FLOPs saved. The dotted line is the dense prefill (90.16\%), and the shaded band marks the configuration we use.}
    \label{fig:wam-cache-sweeps}
    \vspace{8pt}
    \captionof{table}{Real-world results on the AIRBOT Play arm (20 trials per task).}
    \label{tab:wam-cache-real}
    \small
    \setlength{\tabcolsep}{6pt}
    \begin{tabular}{l S[table-format=2.1] S[table-format=2.1] S[table-format=2.1] S[table-format=1.3] S[table-format=-2.1]}
        \toprule
        & \multicolumn{3}{c}{SR (\%) $\uparrow$} & {FLOPs} & {$\Delta$FLOPs} \\
        \cmidrule(lr){2-4}
        Method & {Carrot in Bowl} & {Stack Cubes} & {Avg.} & {(T) $\downarrow$} & {(\%)} \\
        \midrule
        Fast-WAM (dense prefill) & 90.0 & 80.0 & 85.0 & 1.123 & 0.0 \\
        \textbf{+ WAM-Cache (ours)} & \bfseries 90.0 & \bfseries 75.0 & \bfseries 82.5 & \bfseries 0.672 & \bfseries -40.2 \\
        \bottomrule
    \end{tabular}
\end{figure}

\textbf{Hyperparameter sensitivity.}
Figure~\ref{fig:wam-cache-sweeps} sweeps the age bound $G$ and the attention budget $b_a$ on clean RoboTwin~2.0.
A larger $G$ keeps tokens cached longer, which saves FLOPs but lets error accumulate.
Moving from $G=1$ to $G=2$ saves 10.2 points of FLOPs for only 0.6 points of success.
Relative to $G=2$, every larger bound costs more than a quarter of a point of success per point of FLOPs saved, so we use $G=2$.
Success rises steeply with $b_a$ up to 0.4 and then saturates.
At $b_a=0.6$, WAM-Cache comes within 0.1 points of the dense prefill while still removing 28.2\% of the prefill FLOPs.
The surprise budget $b_s$ only needs to cover moving tokens, so we keep it small.
Under randomization, where more tokens move, raising $b_s$ from 0.2 to 0.3 gains 1.2 points of success for 2.3 points of FLOPs (Table~\ref{tab:wam-cache-main}).
We keep $G=2$ on every benchmark and on the real robot without retuning.

\subsection{Real-World Experiments}
\label{sec:real-exp}
We further evaluate WAM-Cache on two real-world tasks: placing a carrot in a bowl, which requires grasping an irregular object, and stacking a red cube on a blue cube, which requires precise alignment under contact.
We collect 100 demonstrations per task via teleoperation on the AIRBOT Play platform and train Fast-WAM on them for 30k steps.
We then apply WAM-Cache with the default clean budgets $(b_a,b_s)=(0.4,0.1)$ and $G=2$ without retuning.
Observations come from an external camera and a wrist camera.
Object placements are randomized in both the demonstrations and the 20 evaluation trials per task, and each trial is scored as a binary success without human intervention.
As shown in Table~\ref{tab:wam-cache-real}, WAM-Cache removes 40.2\% of the prefill FLOPs, matches the dense model on carrot placement, and loses a single trial on cube stacking.
The budgets chosen in simulation therefore work on the real robot without retuning (setup and rollouts in Appendix~\ref{app:real-setup}).

\section{Conclusion}
\label{sec:conclusion}

In this work, we presented \textbf{WAM-Cache}, to the best of our knowledge, the first training-free key--value caching framework that targets the dominant computational bottleneck of World Action Models: the video DiT prefill.
We show that visual change or representation drift alone is an insufficient refresh signal: even an oracle with access to ground-truth KV drift plateaus far below the dense policy.
Instead, downstream action accuracy is governed by \emph{where the policy attends} rather than \emph{what moved}.
WAM-Cache therefore combines the action expert's cross-attention with visual latent surprise and adds an age bound on every token, which keeps KV reuse staleness-bounded without retraining.
Across RoboTwin~2.0, LIBERO, and real-world manipulation on an AIRBOT Play arm, WAM-Cache cuts video DiT prefill FLOPs by 32--42\% while staying within 0.7--1.8 points of the dense policy in simulation and 2.5 points on a real robot.
These FLOPs savings yield CUDA-latency speedups that grow as the visual context scales.
However, WAM-Cache still leaves room for improvement.
At small visual token counts, the measured speedup trails the FLOPs reduction because fixed GPU execution and kernel-launch overheads take a constant share of each layer.
This gap narrows as the number of visual tokens grows, which is the regime in which the prefill dominates.

\clearpage
\bibliography{references}

\clearpage
\appendix
\setlength{\belowcaptionskip}{6pt}
\section{WAM-Cache Algorithm}
\label{app:algorithm}

Algorithm~\ref{alg:wam-cache} gives the per-chunk procedure of Section~\ref{sec:method}.

\begin{algorithm}[h]
    \caption{WAM-Cache: one inference chunk $t>0$}
    \label{alg:wam-cache}
    \begin{algorithmic}[1]
        \REQUIRE observation $o_t$; retained $\{\mathbf{K}^l,\mathbf{V}^l\}_{l=1}^{L}$, references $\{\mathbf{z}_{\tau_i}[i]\}$, counters $\{g_i\}$, previous attention $\{\mathbf{A}^l_{t-1}\}$; budgets $b_s,b_a$; age bound $G$
        \STATE $\mathbf{z}_t\leftarrow\operatorname{Enc}(o_t)$
        \STATE $\mathcal{S}_t\leftarrow\operatorname{TopK}(\mathbf{s}_t,b_s)$, Eq.~(\ref{eq:surprise})
        \STATE $\mathcal{P}_t\leftarrow\operatorname{TopK}(\mathbf{p}_t,b_a)$, Eq.~(\ref{eq:action-mass})
        \STATE $\mathcal{A}_t\leftarrow\{i:g_i\ge G\}$ if $G>0$, else $\varnothing$, Eq.~(\ref{eq:wam-age-set})
        \STATE $\mathcal{R}_t\leftarrow\mathcal{S}_t\cup\mathcal{P}_t\cup\mathcal{A}_t$, Eq.~(\ref{eq:wam-refresh-set})
        \FOR[video DiT prefill on $\mathcal{R}_t$]{$l=1$ \TO $L$}
            \STATE recompute $\mathbf{H}^l_t[\mathcal{R}_t]$; update $\mathbf{K}^l[\mathcal{R}_t],\mathbf{V}^l[\mathcal{R}_t]$, Eq.~(\ref{eq:wam-partial-kv})
            \STATE queries of $\mathcal{R}_t$ attend over full $\mathbf{K}^l,\mathbf{V}^l$ and the cached text KV
        \ENDFOR
        \STATE run action expert for $N$ denoising steps over prefilled $\{\mathbf{K}^l,\mathbf{V}^l\}$; record $\{\mathbf{A}^l_t\}$ at step 1
        \STATE $\tau_i\leftarrow t$, $g_i\leftarrow0$ for $i\in\mathcal{R}_t$; $g_i\leftarrow g_i+1$ otherwise, Eq.~(\ref{eq:wam-age-update})
        \ENSURE action chunk
    \end{algorithmic}
\end{algorithm}

\section{Cost Analysis}
\label{app:cost}

This appendix derives the cost statements of Section~\ref{sec:method-cost}.

\textbf{Prefill FLOPs.}
We count a multiply--accumulate as two FLOPs.
Eq.~(\ref{eq:wam-dense-cost}) omits normalization and modulation terms of order $S_v d$.
The self-attention term covers the query, key, value, and output projections and the two $S_v\times S_v$ attention products.
The cross-attention term covers the query and output projections and the two $S_v\times S_c$ products.
The MLP term covers its two linear layers.
The text keys and values of $\mathcal{F}_{\mathrm{CA}}$ do not appear.
They depend only on the instruction, which is constant within an episode, so they are projected once at the first chunk and reused by every later chunk, for the dense model and for WAM-Cache alike.

WAM-Cache forms queries only for the refreshed tokens, as in Eq.~(\ref{eq:wam-partial-kv}).
Every term of Eq.~(\ref{eq:wam-dense-cost}) is therefore scaled by $\rho_t$, including the two attention products, in which the refreshed queries still attend over all $S_v$ keys.
This gives Eq.~(\ref{eq:wam-cost}), and the same factor applies to the full prefill of $L$ modules.
The relative saving $1-\rho_t$ therefore does not depend on $S_v$, $d$, or $d_f$ and is set by the refresh policy alone.

\textbf{Selection overhead.}
Latent surprise costs $\mathcal{O}(S_v d)$ for the per-token distances of Eq.~(\ref{eq:surprise}).
The action-attention score reuses attention maps that the action expert already computes and aggregates them in $\mathcal{O}(L_a S_a S_v)$.
The age rule costs $\mathcal{O}(S_v)$, and the two top-$k$ selections cost $\mathcal{O}(S_v\log S_v)$.
Each of these is below the $\mathcal{O}(S_v d^2)$ of a single projection in one module, and none is visible in the measured CUDA latency (Section~\ref{sec:sim-latency}).

\textbf{Bounding the recompute ratio.}
Content-based selection alone gives the per-chunk bound $\rho_t\le b_s+b_a$, with equality only when $\mathcal{S}_t$ and $\mathcal{P}_t$ are disjoint.
The age term can exceed this bound in an individual chunk.
In the worst case, synchronized counters force a refresh of every token at once.
The amortized contribution of the age term is bounded, however.
After any refresh, a token's counter resets, so two forced refreshes of the same token are separated by at least $G+1$ chunks.
The time-averaged ratio therefore satisfies
\begin{equation}
    \bar{\rho}\;\le\;b_s+b_a+\frac{1}{G+1}.
    \label{eq:wam-rho-bound}
\end{equation}
The bound tightens to $b_s+b_a+(1-b_s-b_a)/(G+1)$ when the content signals concentrate on a stable subset of tokens, since a token refreshed by content never triggers the age bound.
The measured steady-state ratio of our default configuration, 0.584, lies below both bounds, 0.83 and 0.67, and so does that of every configuration in this paper (Section~\ref{sec:sim-latency}).
It can also fall below $b_s+b_a$, since $\mathcal{S}_t$ and $\mathcal{P}_t$ may overlap.

\section{Implementation Details}
\label{app:base-model}

\textbf{Base model.}
Fast-WAM~\citep{fastwam2026} builds on the Wan2.2-5B video DiT~\citep{wan2025} and adds a 1B-parameter action-expert DiT, while the T5 text encoder and video VAE remain unchanged.
The two branches share attention in a Mixture-of-Transformer structure.
The images of all cameras are spatially concatenated into a single frame before VAE encoding, which yields $S_v$ visual tokens (Section~\ref{sec:prelim}).
On RoboTwin~2.0 the head camera and two wrist cameras form a $384\times320$ (height $\times$ width) frame with $S_v=120$, and on LIBERO the third-person and wrist cameras form a $224\times448$ frame with $S_v=98$.

\textbf{Budget choices.}
Randomization on RoboTwin~2.0 moves distractors and changes appearance, so more tokens are displaced per chunk.
A larger surprise budget ($b_s=0.2$ and $0.3$) keeps task-relevant motion from being crowded out of the surprise set.
On LIBERO the wrist view fills half of the frame, against a third on RoboTwin~2.0.
The gripper and the manipulated object dominate this view, so the policy reads more tokens per chunk.
We therefore raise $b_a$ to $0.5$ and $0.6$ at $b_s=0.1$.

\textbf{Baseline adaptation.}
VLA-Cache~\citep{xu2025vlacache} reuses tokens inside a VLA backbone that both encodes the observation and decodes the action, whereas a WAM separates the two into a video DiT and an action expert.
We therefore apply it to the video DiT prefill.
Tokens whose image patches stay static between replans reuse their cached keys and values, unless they receive high video-to-text attention.
The static threshold is calibrated to match the recompute ratio $\bar\rho$ of WAM-Cache.

\textbf{Evaluation protocol.}
We run 100 episodes per task under each condition on RoboTwin~2.0 and 50 episodes per task on LIBERO, with success averaged over tasks.
Each chunk executes 24 actions before replanning on RoboTwin~2.0 and 10 on LIBERO.

\section{Denoising Steps and Prefill Share}
\label{app:steps}

\textbf{Step sweep.}
Table~\ref{tab:step-sweep} reports the dense Fast-WAM checkpoint for $N\in\{1,2,5,10\}$ action-denoising steps, without caching.
RoboTwin~2.0 follows the protocol of Appendix~\ref{app:base-model} with a fixed seed, and LIBERO uses the released LIBERO checkpoint.
On RoboTwin~2.0, $N=5$ and $N=10$ differ by less than half a point.
$N=2$ is 1.3 and 1.5 points below $N=10$ under clean and randomized scenes, and $N=1$ loses more than 20 points.
On LIBERO the sweep is nearly flat.
$N=2$ is 0.6 points below $N=10$ on average, and even $N=1$ loses only 1.4 points, mostly on Long (94.6 to 91.2) and Goal (96.6 to 95.0).
The step count the action expert needs therefore depends on the benchmark, and $N=2$ is the smallest value that holds on both.
We attribute the failure of $N=1$ on RoboTwin~2.0 to the action distribution.
A single step moves from noise to action along one straight line.
This suffices when the target is nearly unimodal, as in LIBERO's single-arm tasks, but it averages across modes when bimanual tasks admit several valid coordinations.
Unlike the step reduction of LingBot-VA~2.0~\citep{zhang2026lingbotva2}, ours requires no consistency distillation.
Fast-WAM~\mbox{\citep{fastwam2026}} reports $N=10$.
We use $N=2$ by default and measure every success rate against the dense model at the same $N$, so the step reduction and the caching are never conflated.

\begin{table}[H]
    \centering
    \caption{Dense Fast-WAM success rate (\%) versus the number of action-denoising steps $N$ on RoboTwin~2.0 and LIBERO.}
    \label{tab:step-sweep}
    \small
    \setlength{\tabcolsep}{5pt}
    \begin{tabular}{c ccc ccccc}
        \toprule
        & \multicolumn{3}{c}{RoboTwin~2.0} & \multicolumn{5}{c}{LIBERO} \\
        \cmidrule(lr){2-4} \cmidrule(lr){5-9}
        $N$ & Clean & Rand. & Avg. & Spatial & Object & Goal & Long & Avg. \\
        \midrule
        1  & 69.72 & 67.72 & 68.72 & 97.4 & 99.0 & 95.0 & 91.2 & 95.65 \\
        2  & 90.16 & 89.66 & 89.91 & 96.8 & 99.4 & 96.0 & 93.8 & 96.50 \\
        5  & 91.94 & 90.80 & 91.37 & 96.8 & 99.6 & 96.6 & 94.2 & 96.80 \\
        10 & 91.46 & 91.18 & 91.32 & 97.2 & 99.8 & 96.6 & 94.6 & 97.05 \\
        \bottomrule
    \end{tabular}
\end{table}

\textbf{WAM-Cache at $N=10$.}
Table~\ref{tab:wam-cache-n10} repeats the RoboTwin~2.0 comparison of Table~\ref{tab:wam-cache-main} at Fast-WAM's original $N=10$ with the same budgets.
At the first budget, WAM-Cache removes 41.6\% and 39.5\% of the prefill FLOPs under clean and randomized scenes and stays within 0.6 and 1.1 points of the dense model.
These gaps are smaller than the 1.8 and 1.3 points at $N=2$.
At the larger budget, WAM-Cache matches the dense model under clean scenes and comes within 0.2 points under randomized scenes.
The refresh rule therefore carries over to $N=10$ without retuning.

\begin{table}[H]
    \centering
    \caption{WAM-Cache on RoboTwin~2.0 at $N=10$ action-denoising steps (50 tasks $\times$ 100 episodes per cell), with the budget settings of Table~\ref{tab:wam-cache-main}. $\Delta$SR and $\Delta$FLOPs are relative to the dense prefill at $N=10$.}
    \label{tab:wam-cache-n10}
    \small
    \setlength{\tabcolsep}{6pt}
    \begin{tabular}{l l c r r r}
        \toprule
        Scene & Method & $(b_a,b_s)$ & SR (\%) $\uparrow$ & $\Delta$SR (pp) & $\Delta$FLOPs (\%) \\
        \midrule
        \multirow{3}{*}{Clean} & Fast-WAM (dense prefill) & -- & 91.46 & 0.0 & 0.0 \\
        & + WAM-Cache & $(0.4,0.1)$ & 90.82 & $-0.6$ & $-41.6$ \\
        & + WAM-Cache, larger budget & $(0.6,0.1)$ & 91.46 & 0.0 & $-28.2$ \\
        \midrule
        \multirow{3}{*}{Randomized} & Fast-WAM (dense prefill) & -- & 91.18 & 0.0 & 0.0 \\
        & + WAM-Cache & $(0.4,0.2)$ & 90.04 & $-1.1$ & $-39.5$ \\
        & + WAM-Cache, larger budget & $(0.4,0.3)$ & 90.94 & $-0.2$ & $-37.3$ \\
        \bottomrule
    \end{tabular}
\end{table}

\textbf{Prefill share.}
Table~\ref{tab:prefill-share} splits the per-chunk FLOPs at $S_v=120$ into the video DiT prefill, the $N$ action-expert steps, and the VAE encoding of the observation.
The prefill costs 1.053~TFLOPs regardless of $N$, excluding the text keys and values that are projected once per episode.
The action expert costs 0.104~TFLOPs per step.
At $N=2$ the prefill takes 83.5\% of the FLOPs of the two DiTs and 57.5\% of the per-chunk total including the VAE.
At $N=10$ these shares are 50.2\% and 39.5\%.
The prefill therefore dominates the per-chunk cost at both step counts, and more so at the $N=2$ we deploy.
With the 41.6\% prefill reduction of Table~\ref{tab:wam-cache-main}, the two DiTs drop from 1.262 to 0.824~TFLOPs per chunk at $N=2$, a 35\% reduction, or 24\% including the VAE.
Compared with dense Fast-WAM at $N=10$, this is $2.5\times$ fewer DiT FLOPs per chunk.

\begin{table}[H]
    \centering
    \caption{Per-chunk FLOPs (T) of Fast-WAM at $S_v=120$ and the share taken by the video DiT prefill, with and without the VAE encoding.}
    \label{tab:prefill-share}
    \small
    \setlength{\tabcolsep}{6pt}
    \begin{tabular}{c rrr rr}
        \toprule
        & \multicolumn{3}{c}{FLOPs (T)} & \multicolumn{2}{c}{Prefill share} \\
        \cmidrule(lr){2-4} \cmidrule(lr){5-6}
        $N$ & Prefill & Action expert & VAE & DiTs only & incl.\ VAE \\
        \midrule
        2  & 1.053 & 0.209 & 0.571 & 83.5\% & 57.5\% \\
        10 & 1.053 & 1.044 & 0.571 & 50.2\% & 39.5\% \\
        \bottomrule
    \end{tabular}
\end{table}

\section{Real-World Setup and Rollouts}
\label{app:real-setup}

\textbf{Platform.}
The AIRBOT Play is a 6-DoF lightweight arm from DISCOVER Robotics with a 1.5~kg rated payload, a 647~mm reach, $\pm$0.1~mm repeatability, and a 3.78~kg arm mass, fitted with a parallel-jaw gripper.
Joints 1--3 rotate at up to 180$^\circ$/s and joints 4--6 at up to 360$^\circ$/s.
Observations come from an external RGB camera facing the workspace and a wrist-mounted RGB camera, synchronized at 20~Hz, and the two views are concatenated into one frame before VAE encoding.
Actions are streamed to the arm at 20~Hz.

\textbf{Tasks and protocol.}
\emph{Carrot in Bowl} requires grasping a carrot from the table and placing it in a bowl, and succeeds when the carrot rests inside the bowl at the end of the episode.
\emph{Stack Cubes} requires placing a red cube on a blue cube, and succeeds when the red cube rests on the blue cube without falling.
For each task we collect 100 demonstrations via teleoperation with randomized object placements, train Fast-WAM on them for 30k steps, and evaluate 20 trials with placements drawn from the same ranges.
Each trial is scored as a binary success without human intervention.

\begin{figure}[H]
    \centering
    \includegraphics[width=\linewidth]{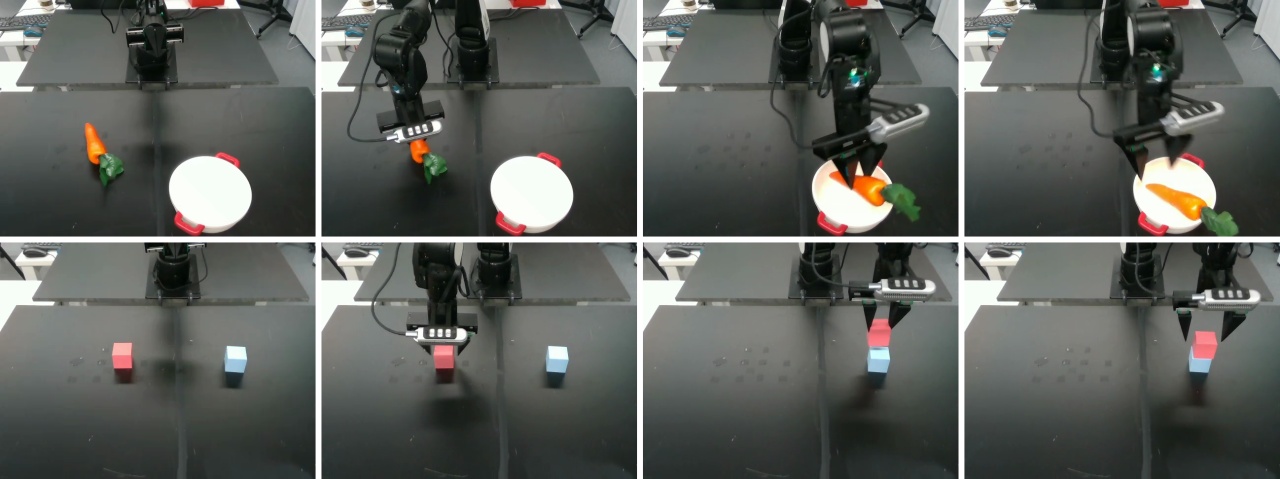}
    \caption{Real-world tasks on the AIRBOT Play arm, viewed from the external camera. Top: picking up a carrot and placing it in a bowl. Bottom: stacking a red cube on a blue cube. Frames run left to right, from the initial scene through grasping and transport to placement.}
    \label{fig:wam-cache-real}
\end{figure}

\textbf{Rollouts.}
Figure~\ref{fig:wam-cache-real} shows one rollout of each task from the external camera.
The arm approaches the object, grasps it, carries it over the target, and releases it, while the rest of the scene stays fixed from frame to frame.

\end{document}